\documentclass[10pt,twocolumn,letterpaper]{article}

\usepackage{wacv}
\usepackage{times}
\usepackage{epsfig}
\usepackage{graphicx}
\usepackage{amsmath}
\usepackage{amssymb}
\usepackage[accsupp]{axessibility}  % Improves PDF readability for those with disabilities.
\usepackage{multirow}
\usepackage{subcaption}

\newcommand{\mf}[1]{\textcolor{black}{#1}}
\def\wacvPaperID{1263} % Enter the WACV Paper ID here

\wacvfinalcopy % *** Uncomment this line for the final submission

\ifwacvfinal
\fi

\ifwacvfinal
\usepackage[breaklinks=true,bookmarks=false]{hyperref}
\else
\usepackage[pagebackref=true,breaklinks=true,colorlinks,bookmarks=false]{hyperref}
\fi

\begin{document}

%%%%%%%%% TITLE
\title{Learnable Multi-level Frequency Decomposition and Hierarchical Attention Mechanism for Generalized Face Presentation Attack Detection \vspace{-6mm}}

\author{Meiling Fang$^{1,2}$, Naser Damer$^{1,2}$, Florian Kirchbuchner$^{1,2}$, Arjan Kuijper$^{1,2}$\\
$^{1}$Fraunhofer Institute for Computer Graphics Research IGD,
Darmstadt, Germany\\
$^{2}$Mathematical and Applied Visual Computing, TU Darmstadt,
Darmstadt, Germany\\
Email: {meiling.fang@igd.fraunhofer.de}
}

\maketitle

\ifwacvfinal
\thispagestyle{empty}
\fi

%%%%%%%%% ABSTRACT
\begin{abstract}
With the increased deployment of face recognition systems in our daily lives, face presentation attack detection (PAD) is attracting much attention and playing a key role in securing face recognition systems. Despite the great performance achieved by the hand-crafted and deep-learning-based methods in intra-dataset evaluations, the performance drops when dealing with unseen scenarios. In this work, we propose a dual-stream convolution neural networks (CNNs) framework. One stream adapts four learnable frequency filters to learn features in the frequency domain, which are less influenced by variations in sensors/illuminations. The other stream leverages the RGB images to complement the features of the frequency domain. Moreover, we propose a hierarchical attention module integration to join the information from the two streams at different stages by considering the nature of deep features in different layers of the CNN. The proposed method is evaluated in the intra-dataset and cross-dataset setups, and the results demonstrate that our proposed approach enhances the generalizability in most experimental setups in comparison to state-of-the-art, including the methods designed explicitly for domain adaption/shift problems. We successfully prove the design of our proposed PAD solution in a step-wise ablation study that involves our proposed learnable frequency decomposition, our hierarchical attention module design, and the used loss function.
Training codes and pre-trained models are publicly released \footnote{\url{https://github.com/meilfang/LMFD-PAD}}.
\end{abstract}

\vspace{-4mm}
%%%%%%%%% BODY TEXT
\section{Introduction}

\begin{figure}[thbp!]
\begin{center}
\includegraphics[width=0.70\linewidth]{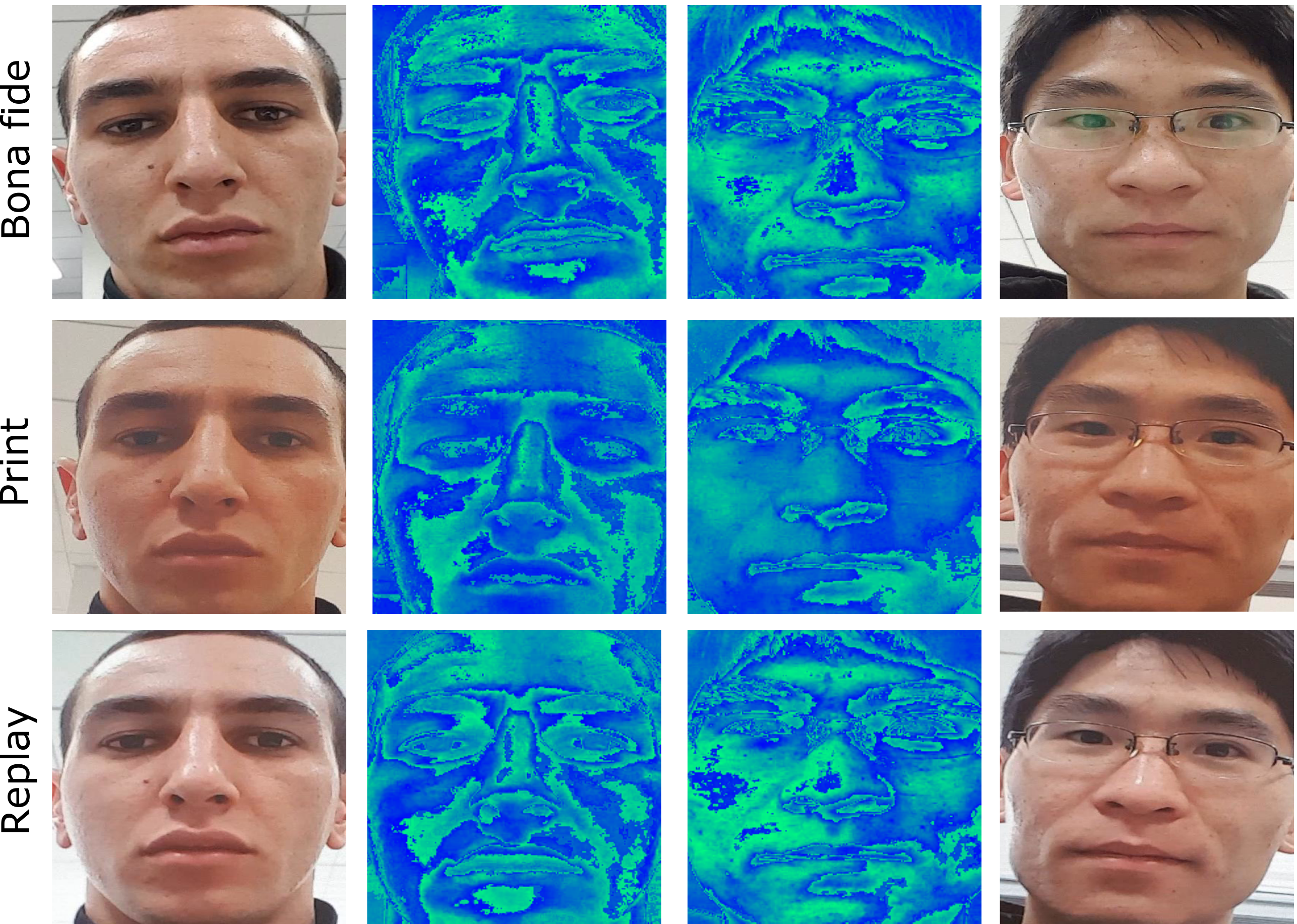}
\end{center}
\caption{The frequency decomposed image components by DCT and inverse DCT obtained according to Equation \ref{eq:dct}, for bona fide (top), print PA (middle), and replay PA (bottom). \mf{The face images are in OULU-NPU dataset \cite{oulu_npu}.} The lighter blue represents the response to high-frequency. It can be observed that the print attack contains relatively less high-frequency information.}
\label{fig:frequency_features_samples}
\vspace{-5mm}
\end{figure}

In recent years, face recognition systems have been widely used in our daily lives for person authentication or access control due to their convenience and remarkable accuracy. However, most existing face recognition systems are vulnerable to Presentation Attacks (PAs). Attackers can use different PAs to impersonate someone or obfuscate their identity. PAs such as print, replay, or 3D mask attacks have been shown to be a serious threat to face recognition systems. Therefore, face Presentation Attack Detection (PAD) plays a critical role in the security of face recognition systems.
PAD methods can be broadly categorized into ones based on hand-crafted features \cite{DBLP:conf/icb/MaattaHP11, DBLP:journals/ejivp/PereiraKAMHPM14, DBLP:journals/tifs/BoulkenafetKH16, LI04, DBLP:conf/bmvc/DamerD16}, and ones based on deep-learning \cite{DBLP:conf/icb/GeorgeM19,DBLP:conf/cvpr/YuZWQ0LZZ20,DBLP:conf/cvpr/LiuJ018,DBLP:journals/tifs/DebJ21,DBLP:journals/pami/YuWQLLZ21}. Hand-crafted based methods utilized traditional texture features such as Local Binary Pattern (LBP) and its extended versions \cite{DBLP:conf/icb/MaattaHP11, DBLP:journals/ejivp/PereiraKAMHPM14, DBLP:journals/tifs/BoulkenafetKH16} that are robust to some variations, e.g., color texture, noise artifacts, in PAs. However, the extracted features may not be discriminative enough between bona fide and attacks. Recent PAD studies \cite{DBLP:conf/cvpr/LiuJ018,DBLP:conf/cvpr/YangLBGGZ0019,DBLP:conf/cvpr/YuZWQ0LZZ20,DBLP:journals/tifs/DebJ21} are competing to boost the performance using Convolution Neural Networks (CNNs) to facilitate more discriminative feature learning. However, CNN-based methods have been a risk of overfitting and thus affect the performance generalization over variations, such as unseen sensors or varied illumination conditions. Considering the characteristics of the hand-crafted and deep learning-based features, it is worth exploring the integration of both features for more discriminative and generalized PAD decisions. 

In addition to widely used LBP features, several studies \cite{LI04,DBLP:journals/tifs/ChenYLWK21,DBLP:conf/mipr/ChenY020} attempted to transform images to the frequency domain. Li \etal \cite{LI04} utilized the dissimilarity in Fourier spectra by considering that less high-frequency components exist in attacks compared to bona fide samples. These hand-crafted features are less relevant to the advanced semantic information like identity information but more relevant to the capture conditions, like displayed screen, used photo, or capture sensors. However, most existing hand-crafted features are extracted by static filters, which might limit the representation capacity and make capturing the relevant patterns harder. A recent study \cite{F3Net_2020} proposed the adaptive partition of images in the frequency domain based on a set of learnable frequency filters to detect face forgery clues. In our work, we adapt several learnable filters to capture the PAs cues. Figure \ref{fig:frequency_features_samples} presents the Discrete Cosine Transform (DCT) based frequency-aware decomposed images. We can observe that the print attacks have less responses to high frequencies (light blue region) compared to the bona fide and replay attacks. Considering the great progress achieved by the deep learning-based methods, we successfully aim at using CNNs to learn subtle differences between bona fide and attacks on both decomposed components in the frequency domain and RGB images in the spatial domain. 

Recently, attention mechanisms were proposed to model the interdependencies between the channel and spatial features on feature maps of CNNs. Woo \etal \cite{cbam} proposed a Convolutional Block Attention Module (CBAM) that can be integrated into any CNN architectures and is end-to-end trainable along with the base CNN. The intermediate feature map is adaptively refined by a combination of channel and spatial wise attention. However, most existing attention-based networks do not consider the nature of features in different layers. The features become more abstract and complex when moving from lower to higher layers in a CNN. The features in the lower layers are relevant to the texture information (e.g., edges), and the features in the higher layers emphasize advanced semantic information. Therefore, simply using a combined channel and spatial attention module may be sub-optimal. In our work, we successfully apply different attention modules according to the nature of the deeply learned features. 

In this work, we aim to integrate learned features from the frequency domain and the spatial domain for better PAD generalization capability. The main contributions of this work are: 
1) We propose a dual-stream PAD solution based on learnable multi-level frequency decomposition (MFD) and our proposed hierarchical attention mechanism (HAM) to capture discriminative and generalize features from both the spatial and frequency domains, namely the LMFD-PAD;
2) An evaluation in both intra-dataset and cross-dataset settings that demonstrates the superiority of our model in cross-dataset PAD when compared to the state-of-the-at, including the PAD methods explicitly targeting domain adaption/shift problem;
3) An ablation study successfully demonstrates the benefits of the proposed LMFD-PAD components, in a step-wise manner, to the cross-dataset PAD performance.

\section{Related work} % Naser DONE % Meiling checked
\label{sec:related_work}
This section reviews the most relevant prior works by focusing on feature-based and deep learning-based face PAD methods, especially those aiming to demonstrate cross-dataset generalizability.

\textbf{Feature-based methods:} Hand-crafted features, such as Local Binary Pattern (LBP) and image distortion, are utilized broadly to detect presentation attacks. For instance, the commonly used LBP projects the faces to a low-dimension representation and has shown good performance on Idiap Replay-Attack dataset \cite{replay_attack}. Boulkenafet \etal \cite{pad_competition} held an IJCB Mobile Face Anti-Spoofing (IJCB-MFAS) competition \cite{pad_competition} carried out on the publicly available OULU-NPU dataset \cite{oulu_npu} in 2017. The goal of the competition was to evaluate the generalizability of PAD algorithms in a mobile environment. The best performing algorithm among all protocols, named GRADIANT, fused color, texture, and motion information from different color spaces. In addition to LBP, transforming face images into the frequency domain was also previously used. Jourabloo \etal \cite{DBLP:conf/eccv/JourablooLL18} used Fast Fourier Transform (FFT) to analyze the spoofing noise. They found that low-frequency features are related to the color distortion and replay artifacts, while high-frequency responses were more obvious on print attacks. Recently, Chen \etal \cite{DBLP:conf/mipr/ChenY020} fused the high and low-frequency features for advanced generalizability of face PAD. In their work, three fixed filters were used to extract the high-frequency information from the input images, and low-frequency features were extracted by Gaussian blur filters. However, the hand-crafted and fixed filters might fail to cover the complete frequency domain, and it is hard to use them to capture features adaptively. Thus, Qian \etal \cite{F3Net_2020} proposed a set of learnable frequency filters for face forgery detection. In our work, we adapt three learnable filters as suggested in \cite{F3Net_2020} and add one more general filter to obtain the frequency-aware decomposed image components, which is complemented by RGB images.

\textbf{Deep learning-based methods:} Deep learning-based methods have been pushing the frontier of face PAD research and have shown significant improvement in PAD performance. George \etal \cite{DBLP:conf/icb/GeorgeM19} proposed a PAD based on pixel-wise and binary supervised (DeepPixBis) training. However, the DeepPixBis method did not generalize well on unseen attacks/sensors scenarios. To further improve the intra-dataset performance and increase the generalization capability, some studies use auxiliary information, e.g., depth \cite{DBLP:conf/cvpr/YuZWQ0LZZ20} and Remote Photoplethysmography (rPPG) signals \cite{DBLP:conf/cvpr/LiuJ018}, for training supervision. For example, Yu \etal \cite{DBLP:journals/pami/YuWQLLZ21} proposed Neural Architecture Search based method for face PAD (NAS-FAS) based on their previous work on Center Difference Convolution Network (CDCN) \cite{DBLP:conf/cvpr/YuZWQ0LZZ20}. They obtained significantly improved results in both intra-dataset and cross-dataset experimental settings. However, the expensive computation cost of NAS must be considered, and the higher error rates in the cross-dataset scenarios suggest that the generalizability is still an open problem. 
Several methods explicitly targeted the domain generalization problem as an inherent domain shift that can be found between different face PAD datasets. Saha \etal \cite{DBLP:conf/cvpr/SahaXKGCPG20} proposed a class-conditional domain discriminator module to generate discriminative bona fide and attack features to tackle the domain shift problem. Most domain generalization face PAD methods \cite{DBLP:conf/cvpr/ShaoLLY19,DBLP:conf/cvpr/LiPWK18,DBLP:conf/aaai/ShaoLY20,DBLP:conf/cvpr/SahaXKGCPG20} performed experiments on four publicly available dataset: Oulu-NPU \cite{oulu_npu}, CASIA-MFSD \cite{casia_fas}, Idiap Replay-Attack \cite{replay_attack}, and MSU-MFSD \cite{msu_mfs}. We follow this cross-dataset setting to compare our method against those state-of-the-art methods later in this work (as reported in Section \ref{ssec:cross-dataset}).

\section{Methodology} % Naser DONE % Meiling checked
In this section, we will provide details of our LMFD PAD solution. We will introduce the multi-level frequency decomposition (MFD), including four learnable frequency filters in Section \ref{ssec:mfd}. Then we introduce the dual-stream network architecture where using a hierarchical attention mechanism to integrate the features learned in frequency and spatial domain in Section \ref{ssec:ham}, and at last present the used loss functions in Section \ref{ssec:loss_func}.

%Please make small introduction that include the LMFD-PAD name and state shortly bu clearly the novel contributions in the methodology

\begin{figure*}[htb!]
\begin{center}
\includegraphics[width=0.80\linewidth]{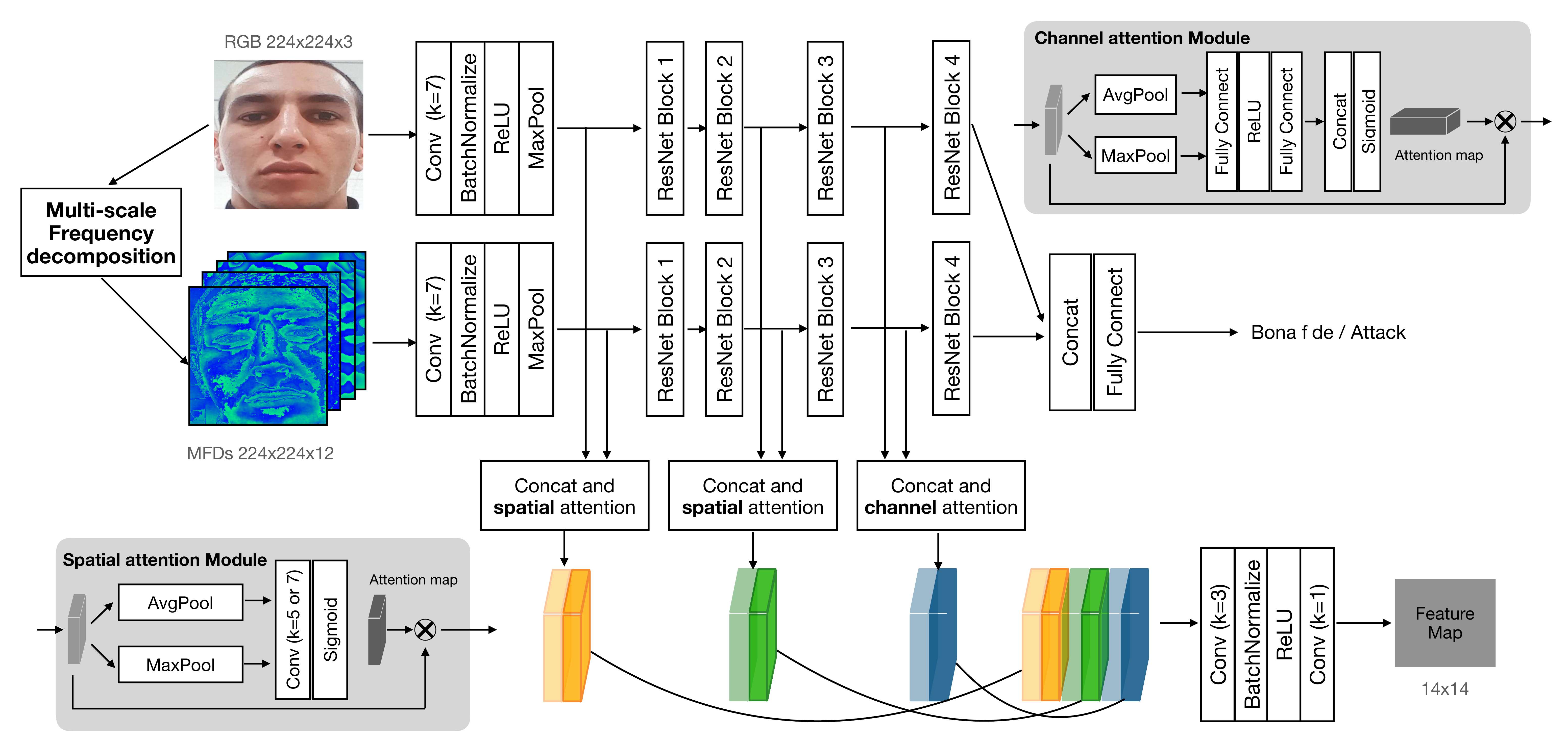}
\end{center}
\vspace{-3mm}
\caption{The overall workflow of our proposed LMFD-PAD solution. Note the utilization of our MFD and HAM (three different channel attention components) components.}
\label{fig:workflow}
\vspace{-4mm}
\end{figure*}

\subsection{Multi-level Frequency Decomposition (MFD)}
\label{ssec:mfd}
Deep-learning based face PAD methods achieved great progress in intra-dataset evaluations. However, the performance normally drops drastically when testing on unseen datasets \cite{DBLP:conf/icb/PurnapatraSBDYM21}. This might be caused by the variations in the attacks and capture environments, such as illuminations and sensors. To address this issue, our proposed LMFD solution decomposes an input face image into different level frequency components. 
Frequency domain analysis is a classical method in image signal processing and has been widely used for general image classification and texture classification tasks \cite{DBLP:conf/mlsp/StuchiAPBFPA17,DBLP:journals/tip/HaleyM99}. Moreover, some face PAD methods attempted \cite{LI04,DBLP:journals/tifs/ChenYLWK21,DBLP:conf/mipr/ChenY020} to transform the images in frequency domain and mine the artifacts cues. The results showed that features in the frequency domain are less sensitive to the variations of the capture environments (e.g., sensors or light conditions). However, most existing frequency-based face PAD methods used filters with fixed weight and maybe sub-optimal for discriminative feature learning.

In our work, we use a set of adaptively learnable frequency filters described in \cite{F3Net_2020} for face forgery detection. First, $N$ manually designed binary base filters $\mathcal{F}_{b} = \{ f_{b}^{i} | 1 \leq i \leq N \}$ partition the frequency domain into low, middle, high frequency bands. The goal of the binary base filters is a roughly equal division of spectrum intp $N$ bands from low frequency to high frequency. Then, $N$ learnable filters $\mathcal{F}_{l} = \{ f_{l}^{i} | 1 \leq i \leq N \}$ are added to such binary base filters. The benefit of such learnable filters is the adaptive selection of the frequency of interest beyond the fixed base filters. Finally, a decomposed image component $\mathrm{C}_{i}$ of an input image $x$ can be computed following the equation:
\begin{equation}
\mathrm{C}_{i} = \mathcal{D}^{-1} \{ \mathcal{D}(x) \odot [f_{b}^{i} + \sigma (f_{l}^{i})] \}, i =\{1, ..., N \},
\label{eq:dct}
\end{equation}
where $\mathcal{D}$ is DCT, $\mathcal{D}^{-1}$ is inversed DCT, and $\odot$ is the element-wise product. The $\sigma (f) = \frac{1-exp(-f)}{1+exp(-f)}$ is used to normalize the value of $f$ between $-1$ and $+1$. 

In our case, $N$ is set to 4 to obtain explicitly divided frequency domain of low, middle, and high-frequency bands and the complementary full frequency band. Three bands are chosen as described in \cite{F3Net_2020}: 1) the low frequency band $f_{base}^{1}$ is the first 1/16 of the entire spectrum, 2) the middle frequency band $f_{base}^{2}$ is between 1/16 and 1/8 of the entire spectrum, 3) the high frequency band $f_{base}^{3}$ is between 1/8 and 7/8 of the entire spectrum. However, the partitioned frequencies may not be sufficient to obtain subtle cues between bona fide and attacks. Therefore, we add one additional learnable filter $f_{base}^{4}$ where the frequency band is the entire spectrum. Moreover, we also keep the input RGB image to provide more visual information and complementary to frequency domain information (as shown in Figure \ref{fig:workflow}). 

In the experiments, face detection is firstly performed on the input image by MTCNN framework \cite{DBLP:journals/spl/ZhangZLQ16}. Then, the detected RGB face image is resized to $224 \times 224 \times 3$ pixels. According to the Equation \ref{eq:dct}, four obtained components are stacked along the channel axis, i.e, the size of a stacked decomposition is $224 \times 224 \times 12$. Then, we utilize dual-stream (RGB and MFD) networks to extract different features in a face image (see Figure \ref{fig:workflow}). In our work, we use the ResNet-50 \cite{DBLP:conf/cvpr/HeZRS16} as our backbone network. 

\subsection{Hierarchical Attention Mechanism (HAM)}
\label{ssec:ham}
So far, we use the dual-stream to learn discriminate features in parallel, which may be sub-optimal for a final PAD decision. To enhance that, we propose our hierarchical attention mechanism (HAM) to integrate features from the frequency domain and semantic image domain and to utilize the features from different layers in the dual-stream. 

This HAM is inspired by Convolutional Block Attention Mechanism (CBAM) \cite{cbam}, which proposed channel and spatial attention blocks for the general computer vision task, and Attention Pixel-wise Binary Supervision (A-PBS) method \cite{DBLP:conf/icb/FangDBKK21}, which employed and fused spatial attention features from multi-layers for the iris PAD task. The CBAM \cite{cbam} consisting of the channel, and distinctive spatial sub-modules can be added into networks according to the custom design needs and showed improvements in classification and detection performance with various neural architectures. A-PBS method \cite{DBLP:conf/icb/FangDBKK21} adopted only spatial attention module (i.e., no channel attention module) aiming to locate the most informative region in an RGB eye image, where might contribute most to a PAD decision. However, in our MFD stream, we have multi-level frequency features, and the weights of filters are adaptively learning while the model is training. The high-frequency component emphasizes features like edges and texture information, while the low-frequency component is related to the spatial distribution of the color gamut. Therefore, channel attention is additionally applied in our framework.

Figure \ref{fig:workflow} shows that spatial attention modules are inserted after the first convolution block and the second ResNet block, respectively, while a channel attention module is added following the third ResNet block. The reason for such attention modules arrangement is based on the nature of the features extracted from different layers. The features from lower to higher layers become more abstract and complex. More specifically, the features in the lower layers are related to the appearance and texture cues, and the features in the higher layers might reveal the semantic content information. Consequently, we perform a spatial attention module on a fused feature in lower layers to focus on texture details like the edge. Then, a channel attention module is added after the third ResNet block to learn the advanced semantic features. To be consistent with the observation on the nature of features in different layers, the size of the convolutional kernel is $7 \times 7$ in the first spatial attention module and $5 \times 5$ in the second spatial attention module, as the smaller convolutional kernel is more suitable for locating the small-scale texture cues. Finally, the attentive features are fused to preserve richer patterns. Moreover, we use pixel-wise and binary supervision to train the dual-stream networks as suggested in \cite{DBLP:conf/icb/GeorgeM19} where the intermediate feature map can be considered as the scores generated from the patches in an image and thus improve the performance. On the one hand, the attentive feature maps from different layers are concatenated and fed to the stacked two convolution layers to output a feature map. The size of the output feature map in our case is $14 \times 14$ for pixel-wise supervision. On the other hand, the features from the last ResNet block in two streams are also concatenated and fed to the fully connected layer for binary supervision. 

\subsection{Loss function}
\label{ssec:loss_func}
Binary Cross Entropy (BCE) loss has proved to perform well when used for pixel-wise and binary supervision \cite{DBLP:conf/icb/GeorgeM19}. Nevertheless, to reduce the sensitivity to outliers in the output feature map, we use the Smooth L1 (SL) function to compute the loss between the output feature map and the ground truth binary mask. For binary supervision, we use the Focal Loss instead of BCE loss because the Focal loss (FL) with a relaxing factor can down-weight easy samples (i.e., samples correctly classified with high confidence) and make the model focus on the hard samples with low classification confidence. The equation for Smooth L1 is shown below as:
\[
\mathcal{L}_{SL} = \dfrac{1}{n} \sum z
\]
\[
\text{where} \quad z = 
\begin{cases}
\dfrac{1}{2}\cdot (y - x)^{2}, & \textit{if} \quad |y - x| < 1 \\
|y-x| - \dfrac{1}{2}, & \textit{otherwise}
\end{cases}
\]
where $n$ is the number of pixels in the output map (14 in our case). $x$ and $y$ refer to the values in the output feature map and the ground truth label, respectively. The equation of Focal loss is:
\[
\mathcal{L}_{FL} = -(1-p_{t})^{\gamma}\log(p_{t})
\]
\[
\text{where} \quad p_{t} = 
\begin{cases}
p, & \textit{if} \quad y=1 \\
1-p, & \textit{otherwise}
\end{cases}
\]
where $p$ is the predicted probability when the ground truth label $y$ is 1 (bona fide in our case) and $\gamma$ is a tunable focusing parameter ($\gamma$ is 2 in our experiments). The overall loss function is given as:
\vspace{-2mm}
\begin{equation}
\mathcal{L}_{overall} = \lambda_{1} \cdot \mathcal{L}_{SL} + \lambda_{2} \cdot \mathcal{L}_{FL}
\label{eq:overall_loss_function}
\vspace{-2mm}
\end{equation}
For exploring the effect of loss functions, we also report the results of BCE loss as used in \cite{DBLP:conf/icb/GeorgeM19} as an ablation study (as shown in Table \ref{tab:ablation_results}).

\vspace{-2mm}
\section{Experiments} % Naser DONE % Meiling checked
%In this section, intra-dataset and cross-dataset experiments are performed to demonstrate the generalization capability of our LMFD-PAD method. In the following, we sequentially introduce the experimental settings, including used datasets, implementation details, and evaluation metrics, then results, ablation study on model components, and t-SNE visualization and analysis.

\subsection{Experimental setting}
\label{sec:implementation}

\textbf{Datasets:} Our method is evaluated on four publicly available face PAD datasets: Oulu-NPU \cite{oulu_npu}, CASIA-MFSD \cite{casia_fas}, Idiap Replay-Attack \cite{replay_attack}, and MSU-MFSD \cite{msu_mfs} under different scenarios.  
Oulu-NPU \cite{oulu_npu} dataset consists of 55 subjects and 5940 videos recorded by six mobile phones. Four protocols are provided to evaluate the generalizability of algorithms. Protocol-1 studies the impact of illumination variations, while Protocol-2 evaluates different attacks created by various instruments. Protocol-3 examines the effect of different capture cameras, and Protocol-4 explores all the challenges above by leave-one-out cross-validation. CASIA-MFSD \cite{casia_fas} includes 50 subjects and 600 videos captured by three different quality cameras. This dataset contains three attack types: warped photo attack, cut photo attack, and video replay attack. Idiap Replay-Attack \cite{replay_attack} contains 50 subjects and 300 videos captured by different sensors and different illumination conditions. Moreover, two types of attacks are included in this dataset: print and replay attacks. MSU-MFSD \cite{msu_mfs} contains 35 subjects and 440 videos captured by two different resolutions of cameras. This dataset also includes two types of attacks, printed photo attacks and replay attacks. The videos in datasets are recorded under different environments with variant cameras and subjects, suitable for cross-dataset domain generalization protocol. Moreover, the subjects in the training set and test set are disjoint in intra-dataset settings.

\textbf{Implementation details:}
%\label{ssec:implementation}
The proposed dual-stream networks are based on ResNet-50 \cite{DBLP:conf/cvpr/HeZRS16} with pre-trained weights on the ImageNet dataset \cite{DBLP:conf/cvpr/DengDSLL009}. The data in all PAD datasets are videos, thus, we sample 10 frames in the average time interval of each video to train and test our method. For each frame, the face is detected and cropped by the MTCNN method \cite{DBLP:journals/spl/ZhangZLQ16} and resized to $224 \times 224 \times 3$ pixels. In the training phase, the SGD optimizer is used with an initial learning rate of 0.001, the momentum of 0.9, and a weight decay of 0.0001. Then, the exponential learning rate scheduler is used with a multiplicative factor of the learning rate decay value ($\gamma$) of 0.995. The ratio of bona fide and attack data is close to 1:1 by simply duplicating the needed images to reduce the effect of biased data. Several data augmentation techniques are used for better generalization ability, including horizontal flip, rotation, cutout, RGB channel shift, and color jitter. To further reduce overfitting, the early stopping technique is utilized with the maximum epochs of 100 and the patience epochs of 15. The batch size in the training phase is 32. In our experiments, the $\lambda_{1}$ in overall loss function \ref{eq:overall_loss_function} is set manually to 1 at the beginning of the training and changed to 100 after five training epochs, while $\lambda_{2}$ is set to 1 in the whole training phase. In the testing phase, a final PAD decision score of a video is a fused score (mean-rule fusion) of all frames.

\textbf{Evaluation metrics:}
We follow the sub-protocols and metrics as defined in the competition \cite{pad_competition} which was performed on the OULU-NPU \cite{oulu_npu} dataset for a fair comparison. The Attack Presentation Classification Error Rate (APCER) \cite{ISO301073} is computed separately for each presentation attack instrument (PAI), e.g., print or replay following the equation:
\vspace{-3mm}
\begin{equation}
ACPER_{PAI} = \dfrac{1}{N_{PAI}} \sum_{i=1}^{N_{PAI}}(1-p_{i}) 
\vspace{-2mm}
\end{equation}
where $N_{PAI}$ is the number of attack samples for a given PAI, $p_{i}$ is the predicted binary label of the $i^{th}$ presentation (0 for bona fide and 1 for attack).
Then, following the OULU-NPU protocol \cite{oulu_npu}, APCER$_{wc}$ is the highest APCER is selected to report the overall performance, i.e., the worst case among all the presentation instruments. The equation is APCER$_{wc}$ = $\max$ (APCER$_{PAI}$) among all PAIs.
Bona Fide Presentation Classification Error Rate (BPCER) \cite{ISO301073} is the proportion of incorrectly classified bona fide samples. Average Classification Error Rate (ACER) is the mean of APCER$_{wc}$ and BPCER.
Moreover, to report the cross-dataset results and to be consistent with previous works 
\cite{DBLP:journals/tifs/BoulkenafetKH16,DBLP:conf/cvpr/LiuJ018,DBLP:journals/pami/YuWQLLZ21,DBLP:conf/cvpr/LiPWK18,DBLP:conf/aaai/ShaoLY20}, we report Half Total Error Rate (HTER) and Area Under the receiver operating Curve (AUC) are used for the cross-dataset domain generalization protocol on OULU-NPU \cite{oulu_npu} , CASIA-MFSD \cite{casia_fas}, Idiap Replay-Attack \cite{replay_attack} and MSU-MFSD \cite{msu_mfs} datasets. The HTER is half of the sum of the APCER and BPCER.

\subsection{Comparison with the State-of-the-Art Methods}
\vspace{-3mm}
\subsubsection{Intra-dataset results on OULU-NPU}
\label{sssec:intra-dataset}
An IJCB-MFAS competition \cite{pad_competition} was carried out on the publicly available OULU-NPU dataset. To assess the generalizability of the face PAD methods, four protocols are provided consisting of cross-environment, cross-PAIs, cross-sensors, cross-all scenarios. For a fair comparison, we strictly follow the definition and evaluation metric of those protocols. 

In this study, we compare our LMFD-PAD method with the best performing method in IJCB-MFAS competition \cite{pad_competition}, GRADIANT. Moreover, we also compare with several recently PAD methods: Auxiliary \cite{DBLP:conf/cvpr/LiuJ018}, FAS-TD \cite{DBLP:journals/corr/abs-1811-05118}, STASN \cite{DBLP:conf/cvpr/YangLBGGZ0019}, DeepPixBis \cite{DBLP:conf/icb/GeorgeM19}, CDCN++ \cite{DBLP:conf/cvpr/YuZWQ0LZZ20}, SSR-FCN \cite{DBLP:journals/tifs/DebJ21}, NAS-FAS \cite{DBLP:journals/pami/YuWQLLZ21} proposed from 2018 to 2021. The results are reported in Table \ref{tab:oulu_results}. \mf{\footnote{The results of state-of-the-art solutions listed in Table \ref{tab:oulu_results} and \ref{tab:cross_test_results} are those reported in their paper.}} The LMFD-PAD achieved ACER values of each protocol are 1.5\%, 2.0\%, 3.4\%, and 3.3\%, respectively. It can be observed that our method obtain competitive results in comparison to state-of-the-art methods. For example, the lowest ACER in the most challenging Protocol-4 is 2.9\% achieved by NAS-FAS \cite{DBLP:journals/pami/YuWQLLZ21}, while our LMFD-PAD ACER value is 3.3\%. This result indicates that our model generalizes well on the cross-test scenarios. Considering that we employ pixel-wise supervision, we can group those PAD methods into three groups based on supervision manner for further comparison. GRADIANT \cite{pad_competition} and STASN \cite{DBLP:conf/cvpr/YangLBGGZ0019} was trained only by binary supervision. DeepPixBis \cite{DBLP:conf/icb/GeorgeM19}, SSR-FCN \cite{DBLP:journals/tifs/DebJ21} and our method utilized the pixel-wise and binary supervision. The left four PAD approaches used depth or/and rPPG supervision. It can be found in Table \ref{tab:oulu_results} that our method possesses improved performance compared to pixel-wise and binary supervised models in most cases but scored below the depth/rPPG supervised networks in some cases. This might drive an extension of our work by generating depth or/and rPPG information to improve the intra-dataset performance. In this case, however, the trade-off between computational resource/time and performance needs to be considered.

\begin{table}[htbp!]
\begin{center}
\resizebox{0.47\textwidth}{!}{
\begin{tabular}{c||c|c|c|c}
\hline
Prot. & Method & APCER$_{wc}$(\%) & BPCER(\%) & ACER(\%) \\ \hline
\multirow{9}{*}{1} & GRADIANT \cite{pad_competition} & 1.3 & 12.5 & 6.9 \\ 
 & Auxiliary \cite{DBLP:conf/cvpr/LiuJ018} & 1.6 & 1.6 & 1.6 \\ 
 & FAS-TD \cite{DBLP:journals/corr/abs-1811-05118} & 2.5 & 0.0 & 1.3 \\
 & STASN \cite{DBLP:conf/cvpr/YangLBGGZ0019} & 1.2 & 2.5 & 1.9 \\ 
 & DeepPixBis \cite{DBLP:conf/icb/GeorgeM19} & 0.8 & 0.0 & 0.4 \\ 
 & CDCN++  \cite{DBLP:conf/cvpr/YuZWQ0LZZ20} & 0.4 & 0.0 & \textbf{0.2} \\
 & SSR-FCN \cite{DBLP:journals/tifs/DebJ21} & 1.5 & 7.7 & 4.6  \\ 
 & NAS-FAS \cite{DBLP:journals/pami/YuWQLLZ21} & 0.4 & 0.0 &  \textbf{0.2}\\ 
 & LMFD-PAD (ours) & 1.4 & 1.6 & 1.5 \\ \hline
\multirow{9}{*}{2} & GRADIANT \cite{pad_competition} & 3.1 & 1.9 & 2.5  \\ 
 & Auxiliary \cite{DBLP:conf/cvpr/LiuJ018} & 2.7 & 2.7 & 2.7 \\ 
 & FAS-TD \cite{DBLP:journals/corr/abs-1811-05118} & 1.7 & 2.0 & 1.9 \\ 
 & STASN \cite{DBLP:conf/cvpr/YangLBGGZ0019} & 4.2 & 0.3 & 2.2 \\ 
 & DeepPixBis \cite{DBLP:conf/icb/GeorgeM19} & 11.4 & 0.6 & 6.0 \\ 
 & CDCN++ \cite{DBLP:conf/cvpr/YuZWQ0LZZ20} & 1.8 & 0.8 & 1.3 \\ 
 & SSR-FCN \cite{DBLP:journals/tifs/DebJ21} & 3.1 & 3.7 & 3.4 \\
 & NAS-FAS \cite{DBLP:journals/pami/YuWQLLZ21} & 1.5 & 0.8 & \textbf{1.2} \\ 
 & LMFD-PAD (ours) & 3.1 & 0.8 & 2.0 \\ \hline
\multirow{9}{*}{3} & GRADIANT \cite{pad_competition} & 2.6 ± 3.9 & 5.0 ± 5.3 & 3.8 ± 2.4  \\
 & Auxiliary \cite{DBLP:conf/cvpr/LiuJ018} & 2.7 ± 1.3 & 3.1 ± 1.7 & 2.9 ± 1.5  \\ 
 & FAS-TD \cite{DBLP:journals/corr/abs-1811-05118} & 5.9 ± 1.9 & 5.9 ± 3.0 & 5.9 ± 1.0 \\ 
 & STASN \cite{DBLP:conf/cvpr/YangLBGGZ0019} & 4.7 ± 3.9 & 0.9 ± 1.2 & 2.8 ± 1.6 \\
 & DeepPixBis \cite{DBLP:conf/icb/GeorgeM19} & 11.7 ± 19.6 & 10.6 ± 14.1 & 11.1 ± 9.4 \\ 
 & CDCN++ \cite{DBLP:conf/cvpr/YuZWQ0LZZ20} & 1.7 ± 1.5 & 2.0 ± 1.2 & 1.8 ± 0.7 \\
 & SSR-FCN \cite{DBLP:journals/tifs/DebJ21} & 2.9 ± 2.1 & 2.7 ± 3.2 & 2.8 ± 2.2 \\ 
 & NAS-FAS \cite{DBLP:journals/pami/YuWQLLZ21} & 2.1 ± 1.3 & 1.4 ± 1.1 & \textbf{1.7 ± 0.6} \\ 
 & LMFD-PAD (ours) & 3.5 ± 3.2 & 3.3 ± 3.2  & 3.4 ± 3.1  \\ \hline
\multirow{9}{*}{4} & GRADIANT \cite{pad_competition} & 5.0 ± 4.5 & 15.0 ± 7.1 & 10.0 ± 5.0 \\ 
 & Auxiliary \cite{DBLP:conf/cvpr/LiuJ018} & 9.3 ± 5.6 & 10.4 ± 6.0 & 9.5 ± 6.0 \\ 
 & FAS-TD \cite{DBLP:journals/corr/abs-1811-05118} & 14.2 ± 8.7 & 4.2 ± 3.8 & 9.2 ± 3.4\\ 
 & STASN \cite{DBLP:conf/cvpr/YangLBGGZ0019} & 6.7 ± 10.6 & 8.3 ± 8.4 & 7.5 ± 4.7 \\ 
 & DeepPixBis \cite{DBLP:conf/icb/GeorgeM19} & 36.7 ± 29.7 & 13.3 ± 14.1 & 25.0 ± 12.7  \\ 
 & CDCN++ \cite{DBLP:conf/cvpr/YuZWQ0LZZ20} & 4.2 ± 3.4 & 5.8 ± 4.9 & 5.0 ± 2.9 \\
 & SSR-FCN \cite{DBLP:journals/tifs/DebJ21} & 8.3 ± 6.8 & 13.3 ± 8.7 & 10.8 ± 5.1 \\
 & NAS-FAS \cite{DBLP:journals/pami/YuWQLLZ21} & 4.2 ± 5.3 & 1.7 ± 2.6 & \textbf{2.9 ± 2.8} \\
 & LMFD-PAD (ours) & 4.5 ± 5.3 & 2.5 ± 4.1  & 3.3 ± 3.1 \\ \hline
\end{tabular}}
\end{center}
\vspace{-3mm}
\caption{The results of the intra-dataset evaluation under the four protocols of the OULU-NPU dataset \cite{oulu_npu}. The bold numbers refer to the lowest ACER in each protocol. Note that our LMFD-PAD achieves competitive performance overall and performs better than most methods that do not use auxiliary information (depth or rPPG) as detailed in Section \ref{sssec:intra-dataset}.}
\label{tab:oulu_results}
\vspace{-5mm}
\end{table}

\vspace{-5mm}
\subsubsection{Cross-dataset results}
\label{ssec:cross-dataset}

\begin{table*}[htbp!]
\begin{center}
\resizebox{0.99\textwidth}{!}{
\begin{tabular}{c||cc|cc|cc|cc}
\hline
\multirow{2}{*}{Method} & \multicolumn{2}{c|}{O\&C\&I → M} & \multicolumn{2}{c|}{O\&M\&I → C} & \multicolumn{2}{c|}{O\&C\&M → I} & \multicolumn{2}{c}{I\&C\&M → O} \\ %\cline{2-9} 
 & HTER(\%) & AUC(\%) & HTER(\%) & AUC(\%) & HTER(\%) & AUC(\%) & HTER(\%) & AUC(\%) \\ \hline \hline
MS LBP \cite{DBLP:conf/icb/MaattaHP11} & 29.76 & 78.50 & 54.28 & 44.98 & 50.30 & 51.64 & 50.29 & 49.31 \\ \hline
Binary CNN \cite{DBLP:conf/eccv/XuLNX14} & 29.25 & 82.87 & 34.88 & 71.94 & 34.47 & 65.88 & 29.61 & 77.54 \\ \hline
IDA \cite{DBLP:journals/tifs/WenHJ15} & 66.67 & 27.86 & 55.17 & 39.05 & 28.35 & 78.25 & 54.20 & 44.59 \\ \hline
Color Texture \cite{DBLP:journals/tifs/BoulkenafetKH16} & 28.09 & 78.47 & 30.58 & 76.89 & 40.40 & 62.78 & 63.59 & 32.71 \\ \hline
LBPTOP \cite{DBLP:journals/ejivp/PereiraKAMHPM14} & 36.90 & 70.80 & 42.60 & 61.05 & 49.45 & 49.54 & 53.15 & 44.09 \\ \hline
Auxiliary(Depth Only) \cite{DBLP:conf/cvpr/LiuJ018} & 22.72 & 85.88 & 33.52 & 73.15 & 29.14 & 71.69 & 30.17 & 77.61 \\ \hline
Auxiliary(All) \cite{DBLP:conf/cvpr/LiuJ018} & - & - & 28.40 & - & 27.60 & - & - & - \\ \hline
NAS-FAS \cite{DBLP:journals/pami/YuWQLLZ21} & 16.85 & 90.42 & 15.21 & 92.64 & \textbf{11.63} & \textbf{96.98} & 13.16 & 94.18 \\ \hline \hline
MMD-AAE \cite{DBLP:conf/cvpr/LiPWK18} & 27.08 & 83.19 & 44.59 & 58.29 & 31.58 & 75.18 & 40.98 & 63.08\\ \hline
MADDG \cite{DBLP:conf/cvpr/ShaoLLY19} & 17.69 & 88.06 & 24.50 & 84.51 & 22.19 & 84.99 & 27.98 & 80.02 \\ \hline
RFMetaFAS \cite{DBLP:conf/aaai/ShaoLY20} & 13.89 & 93.98 & 20.27 & 88.16 & 17.30 & 90.48 & 16.45 & 91.16\\ \hline
CCDD \cite{DBLP:conf/cvpr/SahaXKGCPG20} & 15.42 & 91.13 & 17.42 & 90.12 & 15.87 & 91.72 & 14.72 & 93.08\\ \hline \hline
LMFD-PAD (ours) & \textbf{10.48} & \textbf{94.55} & \textbf{12.50} & \textbf{94.17} & 18.49 & 84.72 & \textbf{12.41} & \textbf{94.95}\\ \hline
\end{tabular}}
\end{center}
\vspace{-3mm}
\caption{The results of the cross-dataset evaluation under different experimental settings on four face PAD datasets. In each setting, three datasets are used for training, and one remaining dataset is used for testing. Our LMFD-PAD method is compared with state-of-the-art face PAD methods reporting on this protocol. Not that four of the state-of-the-art methods MMD-AAE, MADDG, RFMetaFAS, and CCDD are explicitly designed to target the domain shift problem. The bold numbers indicate the lowest HTER and highest AUC in each setting.}
\label{tab:cross_test_results}
\vspace{-4mm}
\end{table*}

\begin{table*}[htbp!]
\begin{center}
\resizebox{0.99\textwidth}{!}{
\begin{tabular}{ccc|cc||cc|cc|cc|cc}
\hline
\multirow{2}{*}{RGB} & \multirow{2}{*}{MFD} & \multirow{2}{*}{HAM} & \multirow{2}{*}{BCE} & \multirow{2}{*}{FL+SL} & \multicolumn{2}{c|}{O\&C\&I → M} & \multicolumn{2}{c|}{O\&M\&I → C} & \multicolumn{2}{c|}{O\&C\&M → I} & \multicolumn{2}{c}{I\&C\&M → O} \\ %\cline{2-9} 
& & & & & HTER(\%) & AUC(\%) & HTER(\%) & AUC(\%) & HTER(\%) & AUC(\%) & HTER(\%) & AUC(\%) \\ \hline \hline
$\surd$ & & &$\surd$ & & 17.14 & 90.47 & 22.12 & 82.10 & 24.62 & 82.28 & 19.47 & 88.16\\ \hline
$\surd$ & $\surd$ & & $\surd$ & & 15.47 & 93.17 & 17.21 & 87.50 & 23.51 & 83.25 & 17.26 & 90.41 \\ \hline
$\surd$ & $\surd$ & $\surd$ & $\surd$ & & 11.19 & 93.39 & 16.83 & 90.62 & 21.42 & 83.92 & 22.27 & 85.98  \\ \hline
$\surd$ & $\surd$ & $\surd$ & & $\surd$ &\textbf{10.48} & \textbf{94.55} & \textbf{12.50} & \textbf{94.17} & \textbf{18.49} & \textbf{84.72} & \textbf{12.41} & \textbf{94.95}\\ \hline
\end{tabular}}
\end{center}
\vspace{-3mm}
\caption{The results of the ablation study on model inputs, components, and loss functions. The ablation study is performed on cross-dataset experimental settings to uncover the components generalizability benefits. One can note that in most experiments, each of the proposed components contributes positively to the cross-dataset PAD performance.}
\label{tab:ablation_results}
\vspace{-4mm}
\end{table*}

In the cross-dataset scenario, four publicly available face PAD datasets: Oulu-NPU \cite{oulu_npu} (O for short), CASIA-MFSD \cite{casia_fas} (C for short), Idiap Replay-Attack \cite{replay_attack} (I for shot), and MSU-MFSD \cite{msu_mfs} (M for short) are used. Three datasets are randomly selected for training and the remained one is used for testing. Specifically, following previous works targeting the domain adaption and generalization capability of face PAD \cite{DBLP:conf/cvpr/LiPWK18,DBLP:conf/cvpr/ShaoLLY19,DBLP:conf/aaai/ShaoLY20,DBLP:conf/cvpr/SahaXKGCPG20}, four settings are performed: O\&C\&I → M, O\&M\&I → C, O\&C\&M → I and I\&C\&M → O. 

In our work, we compare our LMFD-PAD model against eight state-of-the-art face PAD methods including depth/rPPG supervision based Auxiliary \cite{DBLP:conf/cvpr/LiuJ018} and NAS-FAS \cite{DBLP:journals/pami/YuWQLLZ21} which outperformed in intra-testing on OULU-NPU dataset \cite{oulu_npu}. In addition, we also compare our method with four state-of-the-art domain generalization face PAD methods: MMD-AAE \cite{DBLP:conf/cvpr/LiPWK18}, MADDG \cite{DBLP:conf/cvpr/ShaoLLY19}, RFMetaFAS \cite{DBLP:conf/aaai/ShaoLY20}, and CCDD \cite{DBLP:conf/cvpr/SahaXKGCPG20}, which explicitly target the domain shift problem. The results are reported in Table \ref{tab:cross_test_results} where the last four methods are face methods addressing domain shift problems. Our proposed LMFD-PAD method achieves significantly improved performance in three experiment settings. For example, the HTER value of our model is 10.48\% in O\&C\&I → M setting and 12.50\% in O\&M\&I → C and  12.41\% in I\&C\&M → O, while the second-ranking results in those settings are 13.89\%, 15.21\%, and 13.16\%, respectively. Although our LMFD-PAD method is not explicitly designed for the domain shift problem, our method obtains better performance than domain generalization face PAD methods in most cases. The cross-dataset results are consistent with the result in the most challenging intra-dataset Protocol-4 of OULU-NPU dataset \cite{oulu_npu}. We conclude that our method is able to learn more generalized features, which perform well on unseen domains. However, it is still unclear which part of our model benefits the improved results. This question will be answered in the following section by exploring the effect of the MFD, HAM parts, and loss function in an ablation study.

\begin{figure*}[htbp!]
    \centering
    \begin{subfigure}[b]{0.30\linewidth}
     \centering
     \includegraphics[width=\linewidth]{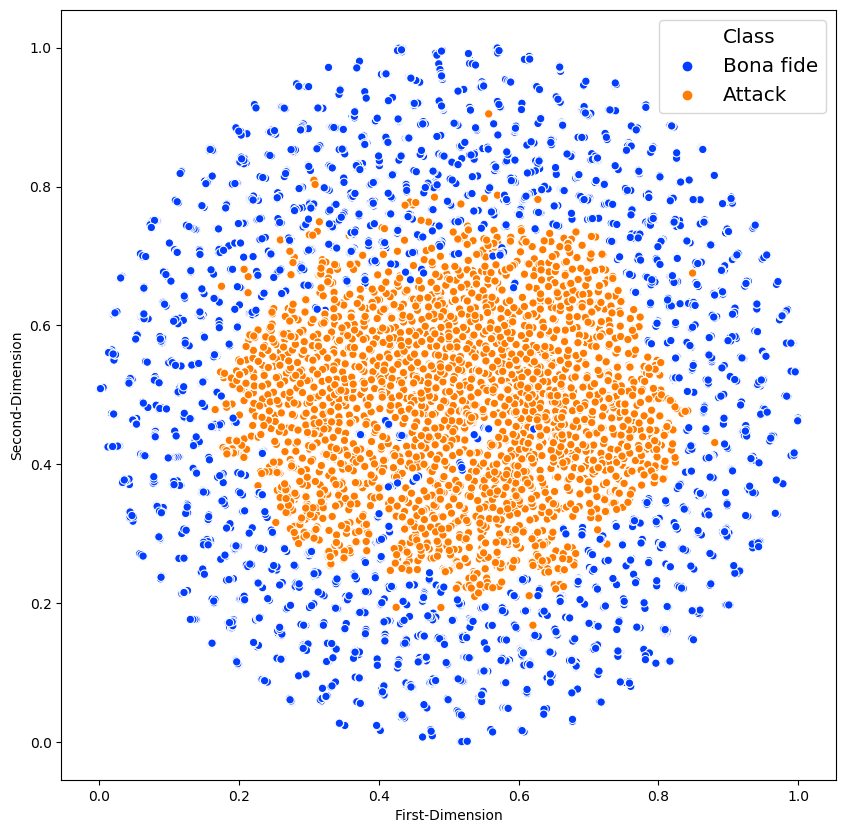}
     \caption{Bona fide and attack.}
     \label{fig:pa}
    \end{subfigure}
    \begin{subfigure}[b]{0.30\linewidth}
     \centering
     \includegraphics[width=\linewidth]{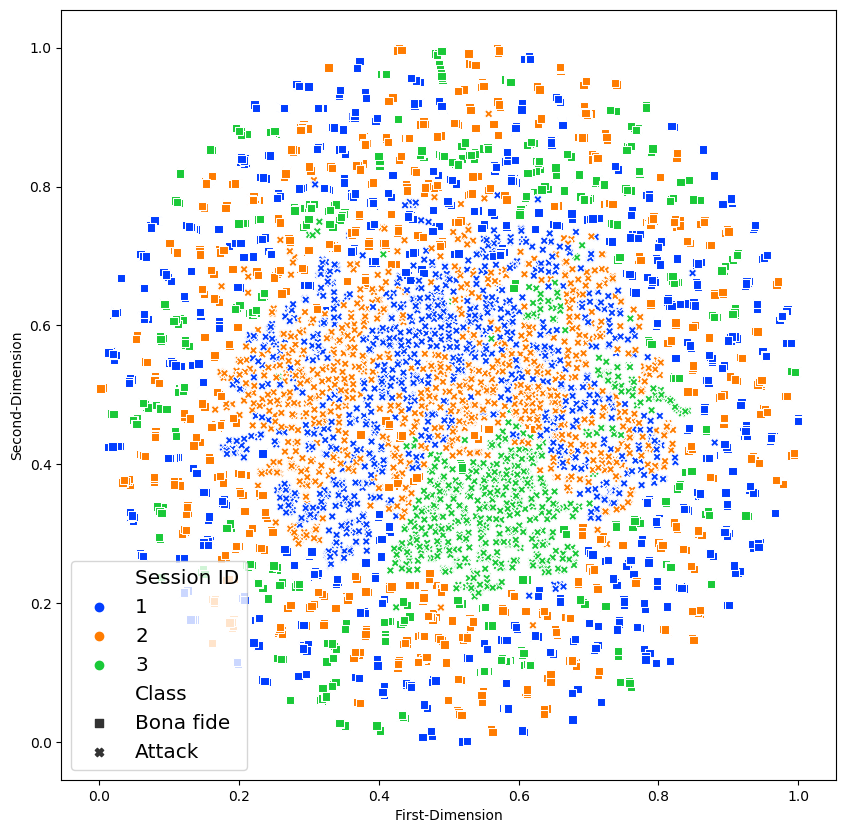}
     \caption{Three different capture scenarios.}
     \label{fig:env}
    \end{subfigure}
    \begin{subfigure}[b]{0.30\linewidth}
     \centering
     \includegraphics[width=\linewidth]{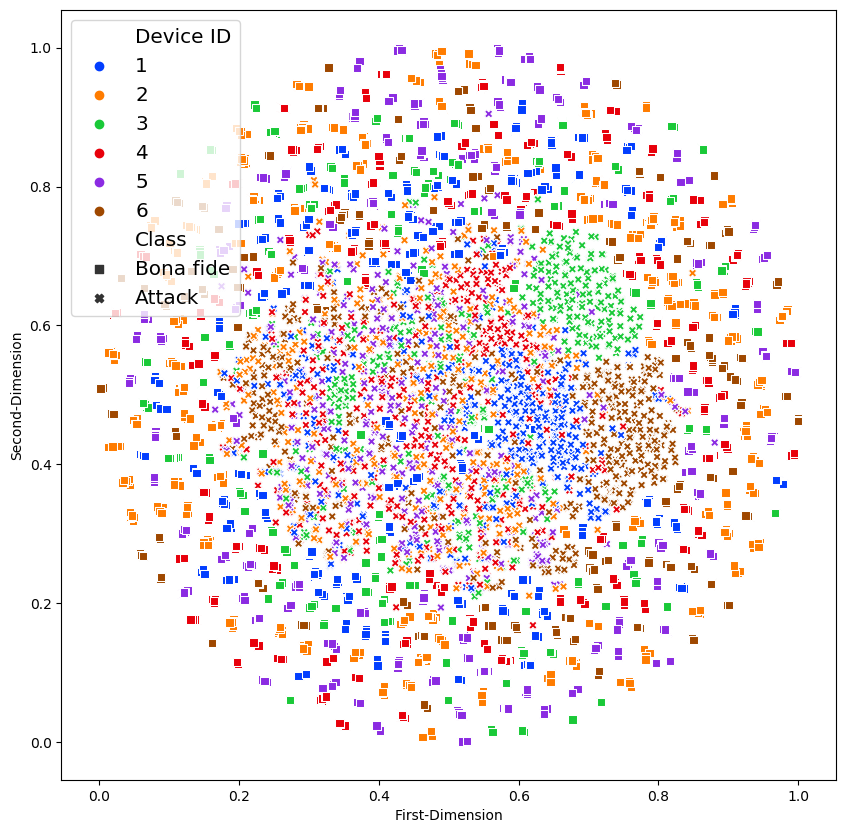}
     \caption{Six different capture devices.}
     \label{fig:phones}
    \end{subfigure}
\vspace{-3mm}
\caption{t-SNE visualization of a cross-dataset setting I\&C\&M → O using our LMFD-PAD embeddings, where the test set is OULU-NPU dataset consisting of three capture environments with different illumination conditions and six mobile devices. The first t-SNE plot represents the two classes: bona fide (blue) and attack (orange). The second and third t-SNE plot indicates three capture scenarios and six capture devices, respectively. In Figures \ref{fig:env} and \ref{fig:phones}, each color corresponds to an environment or device, the signs $\blacksquare$ and $\mathbf{x}$ refers to bona fide and attack, respectively. It is noted that the embeddings from the LMFD-PAD still find a common pattern between the attacks captured under different settings.}
\label{fig:tsne}
\vspace{-4mm}
\end{figure*}

\vspace{-3mm}
\subsection{Ablation study on model components}

So far, the results in Table \ref{tab:oulu_results} and Table \ref{tab:cross_test_results} are obtained by our \textit{full model} including the MFD, HAM and a combined loss function of Focal loss and Smooth L1 loss (Equation \ref{eq:overall_loss_function}). However, the detailed effect of each part is unknown. Therefore, we present an ablation study on model components, and the results are summarized in Table \ref{tab:ablation_results}. This aims at understanding the generalization benefits of each of the proposed components. The training hyper-parameters are the same for all combinations in Table \ref{tab:ablation_results} (training details are described in Section \ref{sec:implementation}). Since we assume that our method is able to learn discriminative and generalize features, the ablation study is demonstrated under the cross-dataset experimental setups on four datasets.

\textbf{Impact of MFD:} To explore the effect of the learnable frequency decomposition, we train a one-stream network using only RGB face images as input and a dual-stream network consisting of RGB and MFD, both solutions are trained by minimizing the BCE loss. The results in Table \ref{tab:ablation_results} shows the improvement by the additional MFD component (the HTER is decreased from 17.14\% to 15.47\% in the O\&C\&I → M setting). A consistent performance enhancement is seen under all the experimental setups in Table \ref{tab:ablation_results}.

\textbf{Impact of HAM:} In contrast to learning the features in the image and the frequency domains in parallel and fusing such features just before the classification layer, we add the HAM component to fuse such features earlier followed by different attention blocks according to the levels of layers, as described in Section \ref{ssec:ham}. The corresponding results are reported in the second row and third row of Table \ref{tab:ablation_results} where it is noticeable that the addition of the proposed HAM did enhance the performance in most experimental settings.   

\textbf{Impact of loss function:} In our LMFD-PAD solution, we use the Focal loss to supervise the binary label prediction and the Smooth L1 to supervise the feature map label prediction instead of the commonly used BCE loss. To explore the effect of such modification, we compare it to using the BCE loss for pixel-wise and binary supervision. The weights of both BCE losses is set to 0.5 as used in \cite{DBLP:conf/icb/GeorgeM19}. 
As presented in Table \ref{tab:ablation_results}, the loss combination used in our LMFD-PAD solution strongly enhances the PAD performance across all the cross-dataset experimental settings.

We conclude that our LMFD-PAD full model boosts the performance generalizability further by adding each of the MFD, HAM, and a combined loss function.
\vspace{-3mm}
\subsection{Visualization and analysis}

In our assumption, the MFD module is able to learn rich generalizable features that adapt well to unseen datasets, especially for unseen sensors or illumination. To further verify this assumption, we use t-SNE \cite{JMLR:v9:vandermaaten08a} plots to visualize deep features in the cross-dataset case I\&C\&M → O. This setting is chosen because the unseen test set is OULU-NPU dataset \cite{oulu_npu} consisting of more variation of environment and capture devices and thus it is better for visualization. The deep features are extracted from the last convolution layer before the classification layer, and then the Principal Component Analysis (PCA) is used to reduce the dimensionality of features to 128-D to reduce the computational cost of the t-SNE. Such features are then projected to 2-D features by t-SNE. Figure \ref{fig:tsne} depicts t-SNE plots on two classes (bona fide and attack), three capture environments, and six capture devices from left to right.
As seen in Figure \ref{fig:pa}, bona fides and attacks \mf{can be considered as coarsely} non-linearly separable. This indicates that our model learns discriminative and generalizes features between bona fides and attacks. In Figure \ref{fig:env}, blue, orange, and green represent three environments of various illuminations. It can be seen that different environments are more obviously clustered in the attack category, while they are clustered more randomly in the bona fide category. A similar observation can be found on different mobile devices in Figure \ref{fig:phones}. These findings suggest that our model is able to mine the general attack artifacts patterns across data capture variety and thus generalizability on unseen datasets is less effect by different sensors or illuminations. This confirms the achieved cross-dataset results in Section \ref{ssec:cross-dataset}.
\vspace{-3mm}
\section{Conclusion}
In this work, we proposed a learnable multi-level frequency decomposition based face PAD method, LMFD-PAD, targeting the generalizability of PAD performance. We employed a dual-stream network architecture. The first stream learns discriminative features in the frequency domain by using learnable frequency filters to obtain frequency decomposed image components, while the other stream uses RGB face images as input to learn features in the spatial domain. Moreover, we proposed the hierarchical attention mechanism to fuse features from both domains at different stages of the network. A spatial attention module is added at the lower layers of the CNN to capture the texture features, and the channel attention module is added at the higher layers of CNN to obtain advanced semantic information. The experiments are demonstrated under intra-dataset and cross-dataset settings. Our LMFD-PAD method achieved comparable results in intra-dataset scenarios. Moreover,  in most cross-dataset cases, our proposed solution outperforms state-of-the-art face PAD methods, including the methods addressing the domain adaption/shift and generalization capability problem. The proposed components of our LMFD-PAD solution are additionally proved in a step-wise ablation study.

\textbf{Acknowledgements:}
This research work has been funded by the German Federal Ministry of Education and Research and the Hessian Ministry of Higher Education, Research, Science and the Arts within their joint support of the National Research Center for Applied Cybersecurity ATHENE.

{\small
\bibliographystyle{ieee_fullname}
\bibliography{egbib}
}

\end{document}